\documentclass[letterpaper, 10 pt, conference]{ieeeconf}  
\IEEEoverridecommandlockouts                              
\usepackage{mathrsfs}
\IEEEoverridecommandlockouts
\usepackage{cite}
\usepackage{amsmath,amssymb,amsfonts}
\usepackage{algorithmic}
\usepackage{algorithm}
\usepackage{graphicx}
\usepackage{textcomp}
\usepackage{float}
\usepackage{hyperref}
\usepackage{xcolor}
\usepackage{geometry}

\def\BibTeX{{\rm B\kern-.05em{\sc i\kern-.025em b}\kern-.08em
    T\kern-.1667em\lower.7ex\hbox{E}\kern-.125emX}}

\title{\LARGE \bf
A Virtual Fencing Framework for Safe and Efficient Collaborative Robotics}

\author{\IEEEauthorblockN{Vineela Reddy Pippera Badguna}
\IEEEauthorblockA{\textit{Dept. of Mechanical \& Aerospace Eng.} \\
\textit{Tandon School of Engineering}\\
\textit{New York University} \\
Brooklyn, NY, USA \\
vp2504@nyu.edu}
\and
\IEEEauthorblockN{Aliasghar Arab}
\IEEEauthorblockA{\textit{Assistant Adjunct Professor} \\
\textit{Tandon School of Engineering}\\
\textit{New York University} \\
Brooklyn, NY, USA \\
aa4764@nyu.edu}
\and
\IEEEauthorblockN{Durga Avinash Kodavalla}
\IEEEauthorblockA{\textit{Dept. of Electrical \& Computer Eng.} \\
\textit{Tandon School of Engineering}\\
\textit{New York University} \\
Brooklyn, NY, USA \\
dk4852@nyu.edu}
}

\author{Vineela Reddy Pippera Badguna$^{1}$, Aliasghar Arab$^{1}$, and Durga Avinash Kodavalla$^{2}$
\thanks{*This work was supported by Mechanical and Aerospace Engineering Department at Tandon School of Engineering in New York University.}
\thanks{$^{1}$Vineela Reddy Pippera Badguna and Aliasghar Arab are with Mechanical and Aerospace Engineering Department and $^{2}$Durga Avinash Kodavalla is with Electrical and Computer Engineering Department at Tandon School of Engineering in New York University, 5 MetroTech Center
Brooklyn, NY 11201   {\tt\small vp2504@nyu.edu, aliasghar.arab@nyu.edu, dk4852@nyu.edu}}
}
\newtheorem{assumption}{Assumption}

\begin{document}

\maketitle
\thispagestyle{empty}
\pagestyle{empty}

\begin{abstract}
Collaborative robots (cobots) increasingly operate alongside humans, demanding robust real-time safeguarding. Current safety standards (e.g., ISO 10218, ANSI/RIA 15.06, ISO/TS 15066) require risk assessments but offer limited guidance for real-time responses. We propose a virtual fencing approach that detects and predicts human motion, ensuring safe cobot operation. Safety and performance tradeoffs are modeled as an optimization problem and solved via sequential quadratic programming. Experimental validation shows that our method minimizes operational pauses while maintaining safety, providing a modular solution for human-robot collaboration.
\end{abstract}

\section{INTRODUCTION}
Cobots, short for collaborative robots, have gained significant traction in various fields, such as manufacturing, assembly, service, education, and healthcare, due to their ability to seamlessly interact with humans while ensuring their physical and mental well-being~\cite{joosse2021making, baek2023uncertainty, hornung2023evaluation}. These robots are specifically designed to operate safely alongside humans, even in close proximity. However, ensuring the safety of people working in the operational workspace of cobots remains a critical challenge. Traditional safety methods, such as physical barriers or complete robot shutdown after human detection, while effective, significantly increase production time~\cite{arab2024safety, 5353980}. Maintaining both human and environmental safety while optimizing cobot performance continues to be an open research frontier, necessitating innovative approaches to enhance safety without compromising efficiency.

The safe integration of industrial robotics and cobots, for human safety, should comply to ISO 10218 globally and ANSI/RIA 15.06 in north America, which outline essential safety requirements for robotic systems. Additionally, ISO/TS 15066 mandates comprehensive risk assessments to mitigate potential hazards in human-robot collaboration, ensuring safer deployment of cobotic systems. Recent advances in real-time human detection have leveraged machine learning pipelines, such as MediaPipe, to halt robot movement upon detecting a person \cite{vu2024real}. Deep-learning-based cobot workstations utilize face recognition and pose detection to adjust robot speed and interactions according to the characteristics of the human operator \cite{8972238}. Such systems pause the cobot's operation when the person is not facing the robot, preventing unintended interactions. While these techniques effectively enhance safety, they introduce productivity challenges due to frequent interruptions in robot operation.

Mixed Reality (MR) technologies have also been explored to improve cobot interaction, providing enhanced control and precision while increasing operator safety\cite{10199911}. Despite its benefits, MR alone does not eliminate the risk of unpredictable collisions, which requires adaptive safety mechanisms. To ensure safe collaboration, the system must be able to autonomously adapt its behavior based on the proximity of human workers, effectively responding to their presence in real time. Dynamic safety systems based on the Speed and Separation paradigm was developed to enable cobots to gradually adjust their speed based on human proximity ~\cite{malm2019dynamic, BYNER2019239, KARAGIANNIS2022102361}. However, the high cost and complexity of sensor integration remain significant challenges, limiting the widespread adaptability of these systems in many collaborative robot applications.

Cobots are equipped with various sensors, including vision, touch, audio, and distance sensors, to prevent potential collisions ~\cite{cherubini2021sensor, BDIWI201465}. Vision-based safety systems are increasingly used to improve human-robot collaboration by enabling real-time human detection and adaptive safety measures \cite{HALME2018111}. Studies highlight the effectiveness of RGB and depth cameras in enforcing safety zones and preventing collisions. A multimodal approach that incorporates depth and thermal cameras has been proposed to enhance human detection, reducing unnecessary stops and slowdowns caused by false positive detections~\cite{9314214, s21217144}. Computer vision-based methods minimize the use of multiple sensors, reducing the complexity of the system \cite{Chebotareva2024}. RGB cameras serve as a cost-effective alternative, offering lower computational requirements while maintaining reliable performance for human-robot collaboration and safety monitoring ~\cite{rodrigues2022collision, 25757, Alenjareghi2024}. RGB camera feeds to automatically control a robot arm when assistance is required~\cite{rekik2024towards}.

The proposed system relies on computer vision-based human detection using an RGB camera. Our proposed method safely minimizes the pause or speed reduction of the cobot based on the determined zones. The system is installed in an UR16e robotic arm and the communication interfaces are built with ROS~2 for a low-latency sensing and control throughput, enabling a modular responsive virtual fence for cobots capable of operating safely and efficiently in a shared workspace.

\begin{figure}[ht]
\centering
\includegraphics[width=0.75\columnwidth]{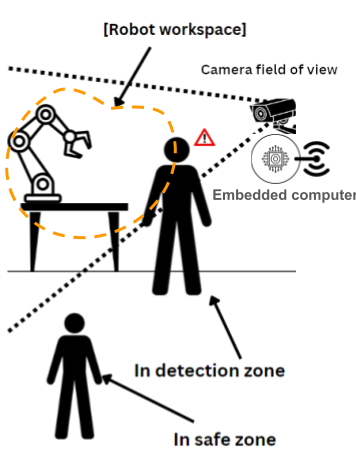}
\caption{Demonstration of a collaborative robot workspace with camera connected to embedded computer to detect human and notify the robot with safeguarding techniques. White dashed lines represent camera field of view.\label{fig:animated-setup}}
\end{figure}

\section{FENCING COLLABORATIVE ROBOTS VIRTUALLY}
Cobots enhance flexibility and efficiency in dynamic industrial settings. However, standard protocols often require shutting them down whenever a person enters their workspace, creating a persistent safety–productivity dilemma. Unpredictable human movement, system latency, and outdated safety methods, such as physical barriers, further impede adaptability and throughput.

A critical challenge arises when human workers enter areas of high risk within the robot’s operational space. Existing safety protocols often adopt a conservative approach, completely stopping the robot’s motion even when the risk is localized. This results in excessive idle time and decreased efficiency. To address this, a more granular and adaptive safety mechanism is needed, one that can dynamically assess risk levels in real time and minimize unnecessary interruptions while maintaining compliance with safety constraints.

In this context, we define safety regions using a structured set of constraints that dictate when and how the cobot must adjust its motion to prevent hazardous interactions. Each safety rule $i \in \mathscr{M}$ defines a safety subset $\mathcal{X}_{\mathrm{s}}^i$ that enforces spatial and dynamic restrictions on robot movements. Here, $\mathscr{M}$ is the set of various rules. The overall safety set $\mathcal{X}_{\mathrm{s}}$ is determined by the intersection of these subsets:
\begin{equation}
\mathcal{X}_{\mathrm{s}}=\bigcap_{i\in \mathscr{M}}{\mathcal{X}_{\mathrm{s}}^i}.
\label{eq:constrSet}
\end{equation}
where each subset $\mathcal{X}_{\mathrm{s}}^i$ is formed by mitigation strategies \(b_i^{(j)}\)$(\mathbf{x}, \mathbf{q}_{\mathrm{e}i})$, which incorporate external data $\mathbf{q}_{\mathrm{e}i}$ from ambient camera to account for dynamic environmental factors:
\begin{equation}
\mathcal{X}_{\mathrm{s}}^i = \left\{\mathbf{x} : b_i^{(j)}(\mathbf{x}, \mathbf{q}_{\mathrm{e}i}) \geq 0, \> \forall{j}, \; 1 \leq j \leq C \right\}.
\label{eq:omega_s}
\end{equation}

\noindent where, $\textbf{x}$ is the state of the robot, $\mathbf{q}_{\mathrm{e}i}$ is the relative distance between the detected person and the center of the critical zone, and $C$ is the number of constraint functions. Here, external data $\mathbf{q}_{\mathrm{e}i}$ include ambient camera feedback in real time, such as human position and proximity estimation. This enables the robot to dynamically redefine its safety constraints based on the evolving environment rather than relying solely on static boundaries. By integrating a camera-based safeguard, the system can continuously monitor human motion and intelligently modulate robot behavior, ensuring safety while reducing unnecessary pauses.

The main objective of this work is to develop an optimized safety framework that minimizes robot downtime by allowing for adaptive risk-aware motion adjustments rather than enforcing blanket shutdowns, except when high-risk operations demand an immediate full stop. Using real-time ambient sensing and dynamically computing the operation method and adjusting safety regions, the proposed approach aims to maintain a balance between safety and operational efficiency in collaborative robotic environments.

\section{HANDLING SAFETY/PERFORMANCE TRADEOFFS} \label{sec-methodology}
This project addresses challenges by formulating a robust control framework that integrates an advanced ambient sensor, real-time control mechanisms, and the person detection algorithm with a time-constrained motion optimization strategy to ensure safe and efficient human-robot interaction. The system continuously monitors a video feed and processes candidate detections generated by a deep neural network. Each candidate detection is represented by a matrix output \(Y \in \mathbb{R}^{N \times 84}\), where for each candidate \(m\) the first four entries correspond to the center-format bounding box, \([c_{xm},\, c_{ym},\, w_m,\, h_m]\), and the subsequent 80 entries represent class logits \([l_{m1},\, l_{m2},\, \ldots,\, l_{m80}]\). Class probabilities are computed via the sigmoid function, that can represented as
\begin{equation}
p_{mn} = \sigma(l_{mn}) = \frac{1}{1+e^{-l_{mn}}},
\end{equation}
and a candidate is considered valid for the “person” class if the maximum probability, defined as
\begin{equation}
s_m = \max_n p_{mn},
\end{equation}
If $s_m$ exceeds a confidence threshold \(\tau\) then the corresponding predicted class, defined as
\begin{equation}
\hat{y}_m = \arg\max_n p_{mn},
\end{equation}
\noindent where, $\hat{y}_m=0$ equals zero. The accepted bounding boxes are then converted to corner format using the relationships, 

\begin{equation}
\label{eq:bound}
\begin{aligned}
x_{1m} &= c_{xm} - \frac{w_m}{2}, \quad y_{1m} = c_{ym} - \frac{h_m}{2}, \\
\quad x_{2m} &= c_{xm} + \frac{w_m}{2}, \quad y_{2m} = c_{ym} + \frac{h_m}{2}.
\end{aligned}
\end{equation}

\begin{assumption}[Bounding Box Classification]
\label{assumption:bounding_boxes}
Each bounding box extracted in Eq.~(\ref{eq:bound}), can be directly classified into \emph{safe}, \emph{caution}, or \emph{unsafe} sets based on visual context (e.g., type of object, proximity or danger levels) derived from image data. Consequently, the overall safe set
\(\mathcal{X}_{\mathrm{s}}=\bigcap_{i\in \mathscr{M}}{\mathcal{X}_{\mathrm{s}}^i}\)
can be constructed by aggregating these bounding boxes within a unified risk-based framework.
\end{assumption}

\begin{figure}[ht]
\centering
\includegraphics[width=0.95\columnwidth]{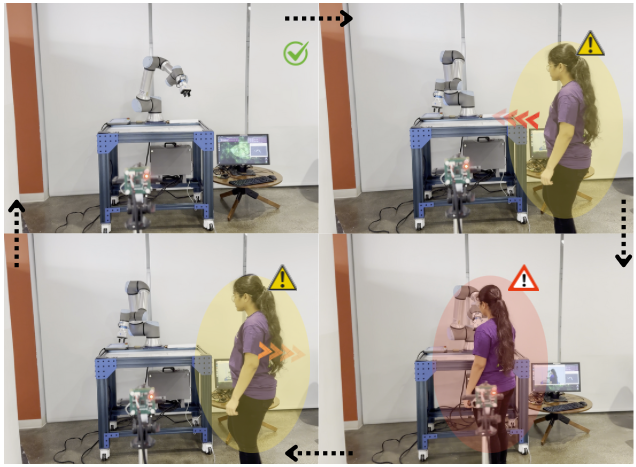}
\caption{The robot operates at normal speed when no person is detected (top left). However, when a person is detected (highlighted in yellow) in the increased attention zone (top right), it slows down. When a person is detected (highlighted in red) in the critical zone (bottom right), the robot stops. Once the person moves back to the increased attention zone (bottom left), the robot resumes slow movement. Once the area is clear, it returns to normal speed.
 \label{fig:animated-zones}}
\end{figure}

The image is vertically divided into three distinct zones according to the overall width \(W\). The central zone, also referred to as the 'critical zone', defined as the interval \(\left[\frac{W}{4},\, \frac{3W}{4}\right]\), corresponds to the robot’s primary workspace, where the presence of humans poses the highest risk as a collision with the robot can occur. The side zones, also referred to as the 'increased attention zone', comprising the intervals \([0,\, \frac{W}{4})\) and \(\left(\frac{3W}{4},\, W\right]\), where a detection prompts a 'slow' command as the human is far from the robot, but the possibility of approaching the robot in the near future can occur. This zoning is deliberately chosen because the robot is only accessible from the front and sides. The rear is protected by a wall and is not accessible to personnel (Fig.~\ref{fig:animated-zones}). In the absence of a detection or if no recent detection is present within a predefined buffer period \(T_{\text{buffer}}\), the system defaults to a 'normal' command.

\subsection{Optimization Problem}
In order to dynamically adjust the robot’s movement in response to the detection state, the framework formulates a constrained optimization problem where the decision variable \(d\) represents the duration of the movement, directly affecting the robot’s speed. The objective is minimizing a quadratic cost function using Sequential Quadratic Programming (SQP) defined as
\begin{equation}
J(d) = \alpha \,(d - d_{\text{desired}})^2 + \beta \,(d - d_{\text{prev}})^2,
\end{equation}
which penalize both deviations from a desired duration \(d_{\text{desired}}\) (set according to the current detection state, such as longer duration for 'slow' commands) and abrupt transitions from the previous duration \(d_{\text{prev}}\). Here, \(\alpha\) and \(\beta\) are weighting factors such that the first term, \(\alpha \,(d - d_{\text{desired}})^2\) penalizes the deviation of \(d\) from the desired duration and the second term, \(\beta \,(d - d_{\text{prev}})^2\) enforces smooth transitions by penalizing sudden changes from previous duration. The solution is subject to safety constraints that ensure,
\begin{equation}
d_{\min} \leq d \leq d_{\max},
\end{equation}
with specific bounds to maintain operation within safe speed limits. In this constrained optimization problem, the Karush-Kuhn-Tucker (KKT) conditions play an important role in ensuring that safety constraints are rigorously enforced. The Lagrange multipliers associated with the inequality constraints serve as indicators of constraint activity. If a constraint is active, that is, the optimal \(d\) is exactly equal to \(d_{\text{min}}\) or \(d_{\text{max}}\), the corresponding multiplier will be positive, indicating that the constraint is binding. In contrast, if the optimal \(d\) is strictly between \(d_{\text{min}}\) and \(d_{\text{max}}\), the multipliers will be zero, indicating that the constraints are inactive at the optimum. The quadratic cost function \(J(d)\) is inherently convex due to its quadratic structure, ensuring a unique global minimum. The gradient of this function,
\begin{equation}
\nabla J(d) = 2\alpha \,(d - d_{\text{desired}}) + 2\beta \,(d - d_{\text{prev}}),
\end{equation}
is linear, and the Hessian is constant,
\begin{equation}
\nabla^2 J(d) = 2\alpha+ 2\beta ,
\end{equation}
which is positive provided that \(\alpha\) and \(\beta\) are positive. This positive definiteness of the Hessian conforms to the strict convexity of the problem, simplifying the optimization landscape and guaranteeing the iterative algorithm. The Sequential Least Squares Programming (SLSQP) algorithm in real time efficiently converges to the unique global minimum, so that the optimized duration \(d\) adapts immediately to changes in the detection command.

When a detection indicates a transition between the 'normal' and 'slow' modes, which means that a person is detected in one of the increased attention zones, the current trajectory is interrupted promptly and a new command is generated based on the updated \(d_{\text{desired}}\). If a person is detected in the critical zone (central), the system issues an immediate 'stop' command to exclude any risk of collision. This integration of detection and optimization guarantees that robot movement continuously adapts to its environment, maintaining both efficiency and safety. The careful division of the field of view into critical and increased attention zones ensures that the system responds appropriately to human proximity while leveraging constraints of the workspace. Such a design, with its rigorous mathematical underpinnings and real-time adaptability, provides a robust and efficient solution for dynamic human-robot interaction scenarios.

\section{EXPERIMENTAL SETUP}

This section details the hardware setup, including the UR16e robotic arm, Arducam IMX477 camera, and Nvidia Jetson Orin Nano for real-time processing, along with data pre-processing for YOLOv8n-based human detection with ONNXRuntime. Experiments evaluate zone-based control with and without SQP-based motion optimization and the immediate stop approach by analyzing robot responses to human presence.

\subsection{Hardware Setup \& Integration}
The experimental framework is designed to evaluate the proposed safety mechanism in a collaborative robotics environment. The setup employs a Universal Robots 16e (UR16e) robotic arm, a high-precision 6-axis manipulator known for its compatibility with the Robot Operating System (ROS)~2, facilitating real-time control and integration with external sensors. The Nvidia Jetson Orin Nano Developer Kit, an AI-optimized and energy-efficient embedded system specialized for robotics and AI applications, serves as the primary computational unit. The system operates on Ubuntu $22.04$ LTS (JetPack 6.0) with ROS~2 Humble version, ensuring reliable and low-latency communication between the robotic arm and peripheral devices.

Visual data for human detection are captured using an Arducam IMX477 high-definition infrared camera, which provides detailed imaging capabilities and integrates efficiently with ROS~2 for streamlined data processing. The Ultralytics YOLOv8n object detection computer vision based PyTorch model is exported as an Open Neural Network Exchange (ONNX) model and is executed via ONNXRuntime, allowing optimized real-time inference on live video streams. To further enhance system responsiveness, a real-time kernel is installed, improving communication between ROS~2 and hardware components, thus ensuring minimal latency.

\begin{figure}[ht]
\centering
\includegraphics[width=0.75\columnwidth]{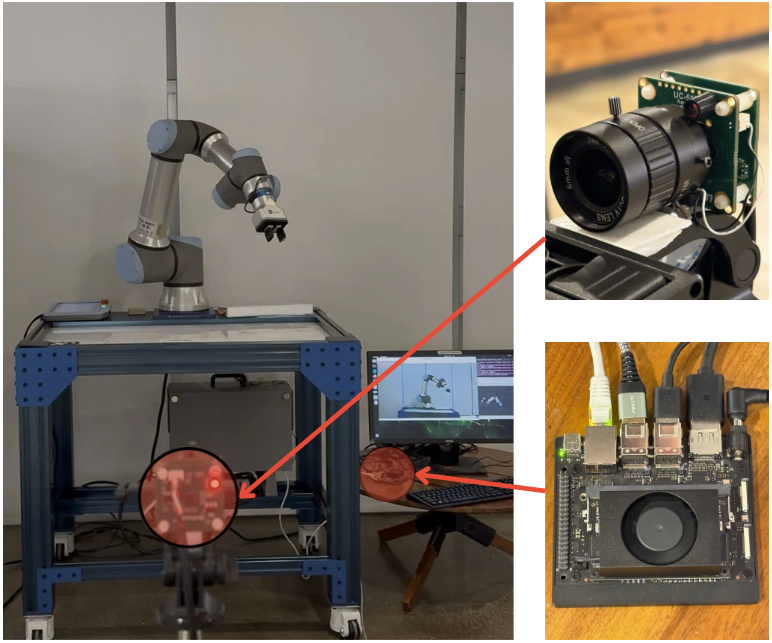}
\caption{Experimental setup featuring a UR16e cobot, integrated with an Arducam IMX477 camera and an embedded AI computational unit.\label{fig:real-setup}}

\end{figure}

The UR16e robotic arm is connected to the computational unit via an Ethernet interface, allowing direct control and seamless execution of safety protocols. In addition, the camera is directly interfaced with the processing unit to provide continuous workspace monitoring (Fig.~\ref{fig:real-setup}). UR ROS2 drivers are installed to establish a communication link between the robotic arm and the computational system, while ROS Visualization (RViz) is used for real-time visualization of robot motion and environmental interactions.

\subsection{Data Pre-processing \& Configuration}
The proposed system integrates a deep learning-based human detection module with a real-time robot control system. A video stream is first acquired from a high-definition camera, after which each frame undergoes a series of pre-processing steps. The frame is resized to a fixed dimension (\(640 \times 640\) pixels) to match the input size of the model, converted from the default BGR color space to RGB, and normalized by scaling the pixel intensities to the interval \([0, 1]\). The processed image is then rearranged into a tensor format with dimensions \((1, C, H, W)\) before being passed to an ONNXRuntime session executing a YOLOv8n model. The model produces an output tensor with dimensions \((1, 84, 8400)\), which follows as described in Section~\ref{sec-methodology}.

The confidence threshold \(\tau\) is set to \(0.65\) and the predefined buffer period \(T_{\text{buffer}}\) is set to 3 seconds for the system to default to the "normal" command when no person is detected in increased attention or critical zones. The weighting factors \(\alpha\) and \(\beta\) are set as \(1.0\) and \(0.85\), respectively, to ensure that the optimized \(d\) reaches \(d_{\text{desired}}\) without sudden changes and smoothly transitions. The safety constraints \([d_{\text{min}}, d_{\text{max}}]\) are set to \([4.0,11.0]\) such that the robot is neither too fast nor too slow in operation, ensuring safety. The \(d_{\text{desired}}\) for the normal, slow and stop commands are set to \(5.0\), \(10.0\), and \(NaN\) (completely halt the robot) seconds, respectively. The real-time display resolution of the live feed is \(1280 \text{ x } 720\) pixels (high definition), along with the inference speed, the detection boundaries, and the zone division information for easy setup and analysis.

\begin{algorithm}[h!]
\caption{Real-Time Person Detection and SQP-Based Motion Smoothening Optimization for Cobot Operations\label{alg:control}}
\begin{algorithmic}[1]
\REQUIRE Video stream \(V\), YOLOv8n ONNX model \(M\), confidence threshold \(\tau\), SQP parameters \(\alpha,\beta,d_{\min},d_{\max}\), initial duration \(d_{\text{prev}}\)
\ENSURE Optimized motion duration \(d^*\) and corresponding robot command
\medskip
\STATE \textbf{Initialization:}
\begin{itemize}
    \item Configure ONNXRuntime with model \(M\).
    \item Initialize video capture from \(V\).
    \item Set default detection flag \(H \leftarrow \text{False}\) and \(d_{\text{prev}}\).
\end{itemize}
\medskip
\WHILE{video stream \(V\) is active}
    \STATE Capture frame \(F\).
    \STATE Preprocess \(F\): resize, convert color (BGR to RGB), normalize, reshape.
    \STATE \textbf{Inference:} Compute output \(O \leftarrow M(F)\).
    \STATE \textbf{Post-processing:}
        \FOR{each candidate detection \(m\) in \(O\)}
            \STATE Compute class probabilities: \(p_{mn} = \sigma(l_{mn}) = \frac{1}{1+e^{-l_{mn}}}\) for \(n=1,\dots,80\).
            \STATE Let \(s_m = \max_n p_{mn}\) and \(\hat{y}_m = \arg\max_n p_{mn}\).
            \IF{\(s_m > \tau\) and \(\hat{y}_m = 0\)}
                \STATE Convert bounding box from center \([c_{xm},\, c_{ym},\, w_m,\, h_m]\) to corner format.
            \ENDIF
        \ENDFOR
    \STATE Determine detection zones
    \STATE Formulate the optimization problem:
    \[
    \min_{d}\; J(d) = \alpha \,(d - d_{\text{desired}})^2 + \beta\,(d - d_{\text{prev}})^2
    \]
    subject to $d_{\min} \le d \le d_{\max}$.
    \STATE Obtain \(d^*\) with SQP.
    \STATE Publish \(d^*\).
    \STATE Update \(d_{\text{prev}} \leftarrow d^*\).
\ENDWHILE
\medskip
\end{algorithmic}
\end{algorithm}

\subsection{Experimentation} \label{subsec-experimentation}
The experimental procedure begins with launching the ROS~2 node that implements Algorithm~\ref{alg:control}, initiating the predefined motion patterns of the robotic arm. A scenario is defined in which the robot performs six cycles (each cycle is a round trip from point A to B and back to A), with an ideal cycle time of \(10\) seconds. The experimental setup involves a human subject who, after the robot completes one cycle, enters the increased attention zone for \(10\) sec, moves into the critical zone for \(10\) sec, returns to the increased attention zone for another \(10\) sec, and finally leaves the frame so that the robot completes the remaining cycles.

The first case focuses on the operational efficiency of two approaches. An experiment uses the proposed model that integrates zone-based detection and SQP-based optimization. The other experiment triggers an immediate stop when detection occurs in any section of the frame and resulting motion only after the area is clear. This comparison underscores the advantage of discriminating between zones, which allows the robot to continue operating safely in slow motion when a person is not that close to the robot but is always ready to stop immediately if too close.

The second case focuses on motion smoothness and is examined in two configurations. One with the proposed framework incorporating SQP optimization for smooth speed transitions, and another using the same detection approach with zones but without the SQP smoothing mechanism. The latter configuration has abrupt transitions between the 'normal' and 'slow' commands. This comparison provides clear evidence of how SQP-based optimization significantly improves the fluidity of the robot’s motion.

\section{RESULTS}
\begin{figure}[ht]
\centering
\includegraphics[width=0.95\columnwidth]{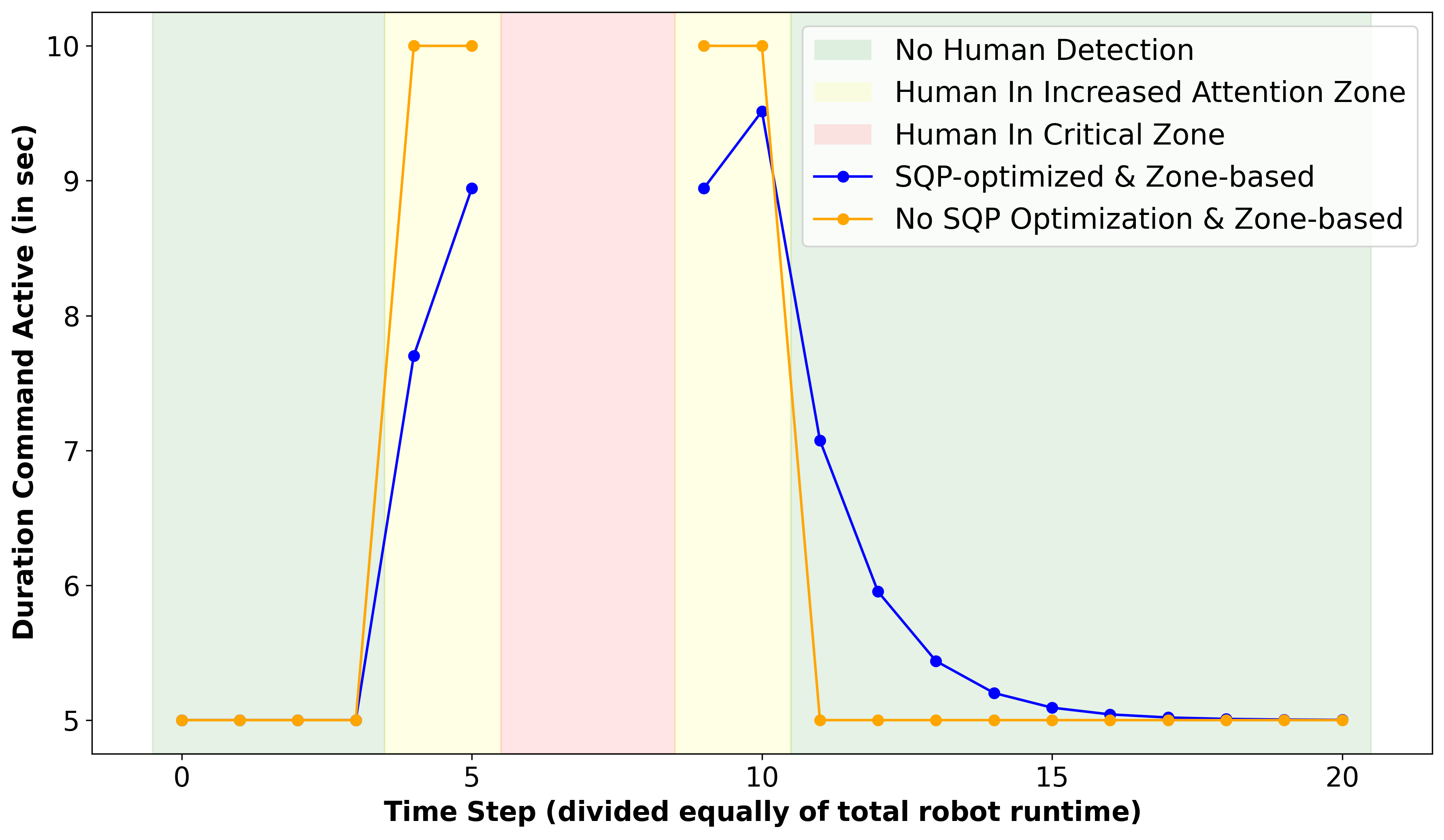}
\caption{The effectiveness of velocity smoothening with and without SQP-based optimization. Zone-based detection is employed in both cases.\label{fig:Results-2}}
\end{figure}

The integration of the proposed experimental setup was systematically tested under the conditions outlined in Subsection~\ref{subsec-experimentation}. In the absence of human detection, the robot completes six cycles in \(60\) seconds, which is considered as the operational efficiency of \(100\%\). The latency of the system is the time taken by the ambient sensor to detect humans in designated zones, process the feed by the detection \& SQP algorithms, issue the corresponding safeguarding command to the robot and execute that command. The collision avoidance rate quantifies the effectiveness of the system in safeguarding collisions between humans and robots. In the first case, zone-based detection with SQP optimization improved operational efficiency compared to the immediate stop approach, reducing unnecessary stops by continuing operation at reduced speed in the increased attention zone (Fig.~\ref{fig:Results-2}) and stopping only when the person is in the critical zone.


\begin{table}[ht]
\centering
\caption{Operational Efficiency (OE), System Latency (SL), and Collision Avoidance Rate (CAR) Comparison Based On Methods Employed \label{table:comparison table}}
\begin{tabular}{p{2.4cm}p{1.1cm}p{1.0cm}p{0.6cm}}
\hline
\textbf{Method} & \textbf{OE} & \textbf{SL} & \textbf{CAR}  \\
\hline
Immediate Stop & 61.8\% & 31.4 ms & 98\%  \\
\hline
Zone-based & 66.7\% & 32.7 ms & 98\% \\
\hline
Zone-based + SQP & 64.5\% & 33.2 ms & 98\% \\
\end{tabular}
\end{table}

In the second case, SQP-based optimization enabled smooth trajectory transitions by minimizing abrupt speed variations (Fig.~\ref{fig:Results-2}). This results in a more adaptive robot motion, which leads to safer and more predictable interactions. In contrast, the system without SQP optimization exhibited sudden speed transitions, leading to jerky movements. Although operational efficiency is higher, the absence of smooth speed transitions poses significant risks, particularly when handling heavy or sensitive payloads, where sudden movements can lead to instability and potential hazards.

The results (Table~\ref{table:comparison table}) validate that the collision avoidance rate is high for all cases, demonstrating a safe environment for smooth human-robot collaborative operations. The proposed zone-based control strategy and SQP-based motion optimization maintain the balance between safety and operational efficiency.

\section{CONCLUSION}
In conclusion, this paper presented a modular safeguarding mechanism designed to enhance collaborative robotic operations in dynamic industrial environments. The proposed approach integrates a real-time human detection and prediction module with a UR16e workstation, allowing a flexible fencing framework. By formulating the safety–performance tradeoff as an optimization problem and solving it with SQP, a zone-based switching control strategy is achieved. Experimental results confirm that the virtual fencing method is cost-effective, exhibits low latency, and can be readily adapted to diverse industrial applications. Furthermore, minimized halts and smooth speed reductions were demonstrated, underscoring their critical role in maintaining operational efficiency.

\section*{ACKNOWLEDGMENT}
\noindent The authors sincerely thank Prof. Katsuo Kurabayashi and Ray Li for their invaluable guidance and encouragement throughout this project. The authors also express their gratitude to the NYU Tandon School of Engineering for providing the resources and facilities.
\begin{small}
\bibliographystyle{IEEEtran}
\bibliography{references}

\begin{thebibliography}{10}
\providecommand{\url}[1]{#1}
\csname url@samestyle\endcsname
\providecommand{\newblock}{\relax}
\providecommand{\bibinfo}[2]{#2}
\providecommand{\BIBentrySTDinterwordspacing}{\spaceskip=0pt\relax}
\providecommand{\BIBentryALTinterwordstretchfactor}{4}
\providecommand{\BIBentryALTinterwordspacing}{\spaceskip=\fontdimen2\font plus
\BIBentryALTinterwordstretchfactor\fontdimen3\font minus \fontdimen4\font\relax}
\providecommand{\BIBforeignlanguage}[2]{{%
\expandafter\ifx\csname l@#1\endcsname\relax
\typeout{** WARNING: IEEEtran.bst: No hyphenation pattern has been}%
\typeout{** loaded for the language `#1'. Using the pattern for}%
\typeout{** the default language instead.}%
\else
\language=\csname l@#1\endcsname
\fi
#2}}
\providecommand{\BIBdecl}{\relax}
\BIBdecl

\bibitem{joosse2021making}
M.~Joosse, M.~Lohse, N.~V. Berkel, A.~Sardar, and V.~Evers, ``Making appearances: How robots should approach people,'' vol.~10, no.~1, pp. 1--24, 2021.

\bibitem{baek2023uncertainty}
W.-J. Baek, C.~Ledermann, and T.~Kr{\"o}ger, ``Uncertainty estimation for safe human-robot collaboration using conservation measures,'' in \emph{17th Int. Conf. Intel. Auton. Sys.}\hskip 1em plus 0.5em minus 0.4em\relax Springer, 2023, pp. 85--102.

\bibitem{hornung2023evaluation}
L.~Hornung, C.~Wurll, and B.~Hein, ``Evaluation of software solutions for risk assessment focusing on human-robot collaboration,'' in \emph{17th Int. Conf. Intel. Auton. Sys.}\hskip 1em plus 0.5em minus 0.4em\relax Springer, 2023, pp. 3--14.

\bibitem{arab2024safety}
A.~Arab, Y.~Mousavi, K.~Yu, and I.~B. Kucukdemiral, ``Safety prioritization by iterative feedback linearization control for collaborative robots,'' in \emph{2024 IEEE Conference on Control Technology and Applications (CCTA)}.\hskip 1em plus 0.5em minus 0.4em\relax IEEE, 2024, pp. 811--816.

\bibitem{5353980}
N.~Pedrocchi, M.~Malosio, and L.~M. Tosatti, ``Safe obstacle avoidance for industrial robot working without fences,'' in \emph{2009 IEEE/RSJ International Conference on Intelligent Robots and Systems}, 2009, pp. 3435--3440.

\bibitem{vu2024real}
T.~Vu, D.-T. Nguyen, D.-T. Nguyen, T.~Vu~Toan, and V.-T. Tran, ``Real-time detection to avoid accidents of interaction between humans and collaborative robot using image processing and machine learning,'' in \emph{2024 IEEE International Conference on Computing, Cybernetics and Cyber-Medical Systems (ICCC)}.\hskip 1em plus 0.5em minus 0.4em\relax IEEE, 2024, pp. 000\,229--000\,232.

\bibitem{8972238}
O.~D. Miguel~Lázaro, W.~M. Mohammed, B.~R. Ferrer, R.~Bejarano, and J.~L. Martinez~Lastra, ``An approach for adapting a cobot workstation to human operator within a deep learning camera,'' in \emph{2019 IEEE 17th International Conference on Industrial Informatics (INDIN)}, vol.~1, 2019, pp. 789--794.

\bibitem{10199911}
R.~R, R.~R. Sathya, S.~V, J.~L. N, and G.~S, ``Enhancing human cobot interaction with mixed reality: A futuristic review,'' in \emph{2023 2nd International Conference on Advancements in Electrical, Electronics, Communication, Computing and Automation (ICAECA)}, 2023, pp. 1--6.

\bibitem{malm2019dynamic}
T.~Malm, T.~Salmi, I.~Marstio, and J.~Montonen, ``Dynamic safety system for collaboration of operators and industrial robots,'' \emph{Open Engineering}, vol.~9, no.~1, pp. 61--71, 2019.

\bibitem{BYNER2019239}
C.~Byner, B.~Matthias, and H.~Ding, ``Dynamic speed and separation monitoring for collaborative robot applications – concepts and performance,'' \emph{Robotics and Computer-Integrated Manufacturing}, vol.~58, pp. 239--252, 2019.

\bibitem{KARAGIANNIS2022102361}
P.~Karagiannis, N.~Kousi, G.~Michalos, K.~Dimoulas, K.~Mparis, D.~Dimosthenopoulos, Önder Tokçalar, T.~Guasch, G.~P. Gerio, and S.~Makris, ``Adaptive speed and separation monitoring based on switching of safety zones for effective human robot collaboration,'' \emph{Robotics and Computer-Integrated Manufacturing}, vol.~77, p. 102361, 2022.

\bibitem{cherubini2021sensor}
A.~Cherubini and D.~Navarro-Alarcon, ``Sensor-based control for collaborative robots: Fundamentals, challenges, and opportunities,'' \emph{Frontiers in Neurorobotics}, vol.~14, p. 576846, 2021.

\bibitem{BDIWI201465}
M.~Bdiwi, ``Integrated sensors system for human safety during cooperating with industrial robots for handing-over and assembling tasks,'' \emph{Procedia CIRP}, vol.~23, pp. 65--70, 2014, 5th CATS 2014 - CIRP Conference on Assembly Technologies and Systems.

\bibitem{HALME2018111}
``Review of vision-based safety systems for human-robot collaboration,'' \emph{Procedia CIRP}, vol.~72, pp. 111--116, 2018, 51st CIRP Conference on Manufacturing Systems.

\bibitem{9314214}
M.~Costanzo, G.~De~Maria, G.~Lettera, and C.~Natale, ``A multimodal approach to human safety in collaborative robotic workcells,'' \emph{IEEE Transactions on Automation Science and Engineering}, vol.~19, no.~2, pp. 1202--1216, 2022.

\bibitem{s21217144}
U.~B. Himmelsbach, T.~M. Wendt, N.~Hangst, P.~Gawron, and L.~Stiglmeier, ``Human–machine differentiation in speed and separation monitoring for improved efficiency in human–robot collaboration,'' \emph{Sensors}, vol.~21, no.~21, 2021.

\bibitem{Chebotareva2024}
E.~Chebotareva, M.~Mustafin, R.~Safin \emph{et~al.}, ``Camera-based safety system for collaborative assembly,'' \emph{Journal of Intelligent Manufacturing}, 2024.

\bibitem{rodrigues2022collision}
I.~R. Rodrigues, G.~Barbosa, A.~O. Filho \emph{et~al.}, ``A new mechanism for collision detection in human–robot collaboration using deep learning techniques,'' \emph{Journal of Control, Automation and Electrical Systems}, vol.~33, pp. 406--418, 2022.

\bibitem{25757}
D.~Gradolewski, D.~Maslowski, D.~Dziak, B.~Jachimczyk, S.~Mundlamuri, C.~Prakash, and W.~Kulesza, ``A distributed computing real-time safety system of collaborative robot,'' \emph{Elektronika ir Elektrotechnika}, vol.~26, pp. 4--14, 04 2020.

\bibitem{Alenjareghi2024}
M.~J. Alenjareghi, S.~Keivanpour, Y.~A. Chinniah, and S.~Jocelyn, ``Computer vision-enabled real-time job hazard analysis for safe human–robot collaboration in disassembly tasks,'' \emph{Journal of Intelligent Manufacturing}, 2024.

\bibitem{rekik2024towards}
K.~Rekik, G.~Silva, A.~Bashir, and R.~M{\"u}ller, ``Towards global awareness in human-robot-collaborative multi-cell assembly system,'' in \emph{2024 IEEE 20th International Conference on Automation Science and Engineering (CASE)}.\hskip 1em plus 0.5em minus 0.4em\relax IEEE, 2024, pp. 5--10.

\end{thebibliography}
\end{small}
\end{document}